%% file: acl_latex.tex
\pdfoutput=1

\documentclass[11pt]{article}

\usepackage{acl}

\usepackage{times}
\usepackage{latexsym}
\usepackage{url}
\usepackage{amsmath}
\usepackage{graphicx}
\usepackage{color}
\usepackage{multirow}
\usepackage{booktabs}
\usepackage{geometry}

\usepackage{subfig}
\usepackage{caption}
\usepackage{tabularx}

\usepackage[T1]{fontenc}

\usepackage[utf8]{inputenc}
\usepackage{microtype}
\usepackage{array}

\makeatletter
\renewcommand{\maketag@@@}[1]{\hbox{\m@th\normalsize\normalfont#1}}%
\makeatother
\title{Multi-Document Scientific Summarization from a Knowledge Graph-Centric View}

\author{ Pancheng Wang,
Shasha Li\footnotemark[2],
 Kunyuan Pang,
 Liangliang He,
 Dong Li, \\
\textbf{Jintao Tang}\footnotemark[2],
\textbf{Ting Wang}\footnotemark[2] \\
School of Computer, National University of Defense Technology, Changsha, China\\
\texttt{\{wangpancheng13, shashali, pangkunyuan10, heliangliang19,}\\
\texttt{lidong1,tangjintao, tingwang\}@nudt.edu.cn}
}

\begin{document}
\maketitle
\renewcommand{\thefootnote}{\fnsymbol{footnote}}
\footnotetext[2]{Corresponding authors.}
\input{data/abstract}
\input{data/introduction}
\input{data/approach}
\input{data/experiments}

\input{data/relatedwork}
\input{data/conclusion}

\section*{Acknowledgements}
This work was supported by the National Key Research and Development Project of China (No. 2021ZD0110700) and Hunan Provincial Natural Science Foundation (Grant Nos. 2022JJ30668). The authors would like to thank the anonynous reviewers for their valuable comments and suggestions to improve this paper.

\bibliography{anthology,custom}
\bibliographystyle{acl_natbib}

\appendix

\section{Baselines}
\label{appen1}

\textbf{LexRank}~\citep{erkan2004lexrank} and \textbf{TextRank}~\citep{mihalcea2004textrank} are two unsupervised graph based extractive summarization models. \textbf{HeterSumGraph}~\cite{wang2020heterogeneous} is a heterogeneous graph-based extractive model with semantic nodes of different granularity.  \textbf{HiMAP}~\citep{fabbri2019multi} expands the pointer-generator network~\citep{see2017get} into a hierarchical network and integrates an MMR module. \textbf{HierSumm}~\citep{liu2019hierarchical} is a
Transformer based model with an attention mechanism
to share information cross-document for abstractive
multi-document summarization. \textbf{MGSum}~\citep{jin2020multi} is a multi-granularity interaction network for abstractive
multi-document summarization. We also consider evaluating on single document summarization models by concatenating multiple papers into a long sequence. \textbf{GraphSum}~\citep{li2020leveraging} is a neural multi-document summarization model that leverages well-known graphs to produce abstractive summaries. We use TF-IDF graph as the input graph.
\begin{figure*}[ht]  
    \centering
   \includegraphics[width=\textwidth]{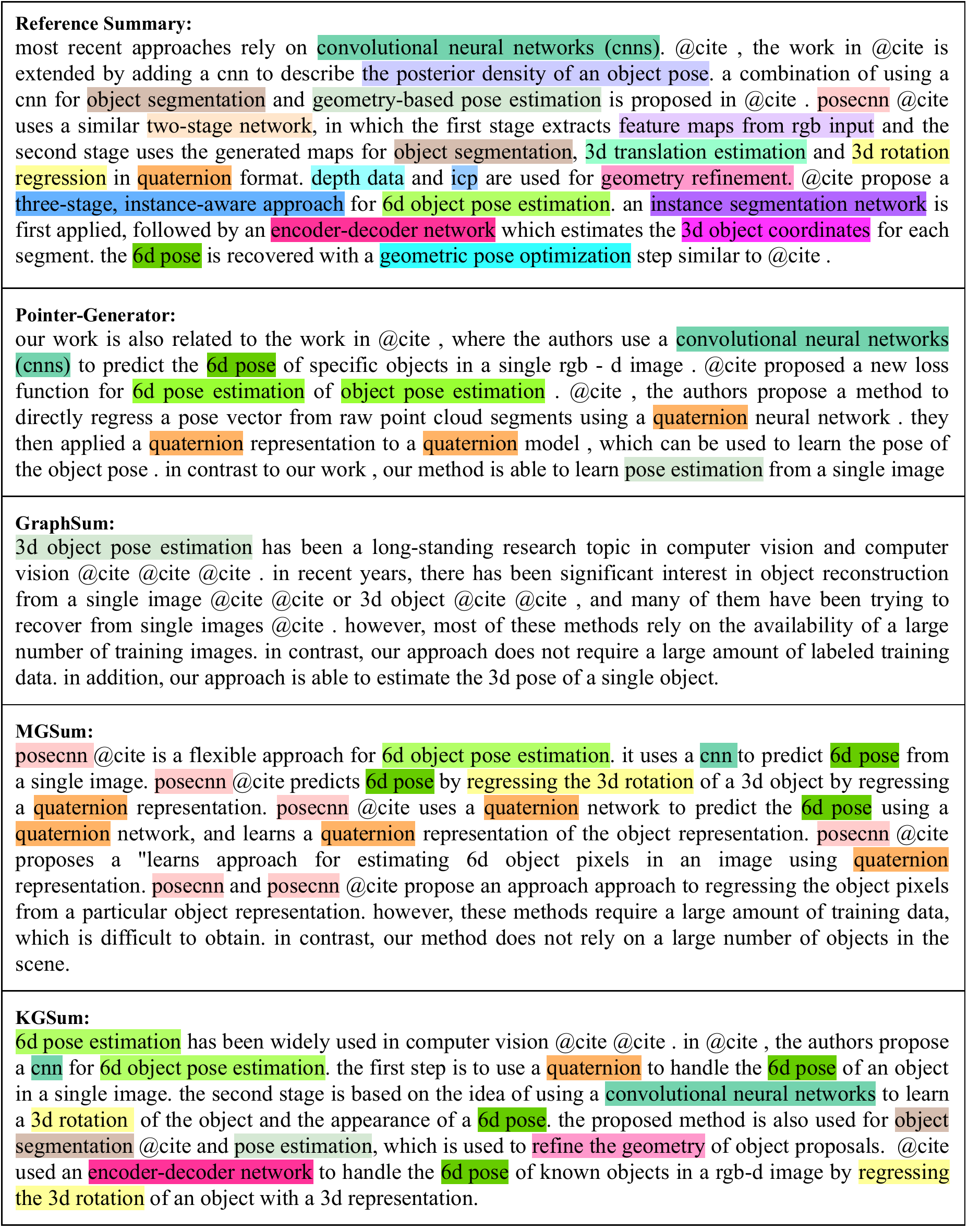}
    
    \caption{Case study from the Multi-Xscience test set. We first highlight the salient contents in the reference summary in different colors. Then the overlapped salient contents of system summaries are annotated in the same colors.} 
    \label{figure9} 
\end{figure*}
\textbf{PEGASUS}~\citep{zhang2020pegasus} is a sequence-to-sequence model with gap-sentences generation as a pre-training objective tailored for abstractive summarization.
\textbf{Pointer-Generator}~\citep{see2017get} is an RNN based model with an attention mechanism and allows the system to copy words from the source via pointing for abstractive summarization.
\textbf{BertABS}~\citep{liu2019text} uses a pretrained BERT~\citep{devlin2019bert} as the encoder for abstractive summarization. We also report the performance of BertABS with an encoder (\textbf{SciBertABS}) pretrained on scientific articles.
\textbf{BART}~\citep{lewis2020bart} is a pretrained text generation model.

\section{Case Study}
In Figure~\ref{figure9}, we present several example summaries to show the generating quality of different models. In the figure, there are five blocks, which are the reference summary (\textbf{Reference Summary}), the system summaries generated by \textbf{Pointer-Generator}, \textbf{GraphSum}, \textbf{MGSum} and our model \textbf{KGSum}. We highlight the salient contents from system summaries that can be find in the reference summary. We could find our model KGSum possesses the most salient contents and the highest overlap with the reference summary. Pointer-Generator and MGSum have the same amount of salient contents, but MGSum contains more repeated information, leading to worse performance. GraphSum contains the least salient contents, resulting in the worst performance.


\end{document}

%% file: data/abstract.tex
\renewcommand{\thefootnote}{\arabic{footnote}}
\begin{abstract}
Multi-Document Scientific Summarization (MDSS) aims to produce coherent and concise summaries for clusters of topic-relevant scientific papers.
This task requires precise understanding of paper content and accurate modeling of cross-paper relationships. Knowledge graphs convey compact and interpretable structured information for documents, which makes them ideal for content modeling and relationship modeling. 
In this paper, we present \textbf{KGSum}\footnote{\url{https://github.com/muguruzawang/KGSum}}, an MDSS model centred on
knowledge graphs during both the encoding and decoding process. Specifically, in the encoding process, two graph-based modules are proposed to incorporate knowledge graph information into paper encoding, while in the decoding process, we propose a two-stage decoder by first generating knowledge graph information of summary in the form of descriptive sentences, followed by generating the final summary. Empirical results show that the proposed architecture brings substantial improvements over baselines on the Multi-Xscience dataset.

\end{abstract}

%% file: data/introduction.tex
\section{Introduction}
\label{intro}
Nowadays, the exponential increasing publication rate of scientific papers makes it difficult and time-consuming for researchers to keep track of the latest advances. \textbf{M}ulti-\textbf{D}ocument \textbf{S}cientific \textbf{S}ummarization (MDSS) is therefore introduced to alleviate this information overload problem by generating succinct and comprehensive summary from clusters of topic-relevant scientific papers \cite{chen2021capturing,shah2021generating}. 

In MDSS, paper content modeling and cross-paper relationship modeling are two main issues. (1) Scientific papers contain complex concepts, technical terms, and abbreviations that convey important information about paper content. However, some previous works~\citep{wang2018neural,jiang2019hsds} treat all text units equally, which inevitably ignore the salient information of some less frequent technical terms and abbreviations. (2) Furthermore, 
there exist intricate relationships between papers in MDSS, such as sequential, parallel, complementary and contradictory~\citep{luu2021explaining}, which play a vital role in  guiding the selection and organization of different contents. The latest work~\cite{chen2021capturing} attempt to capture cross-paper relationships via seq2seq model without considering any links between fine-grained text units. Failure to take into account explicit relationships between papers prevents their model from learning cross-paper relationships effectively.

\begin{figure*}[t]
  \centering
   \includegraphics[width=0.95\textwidth]{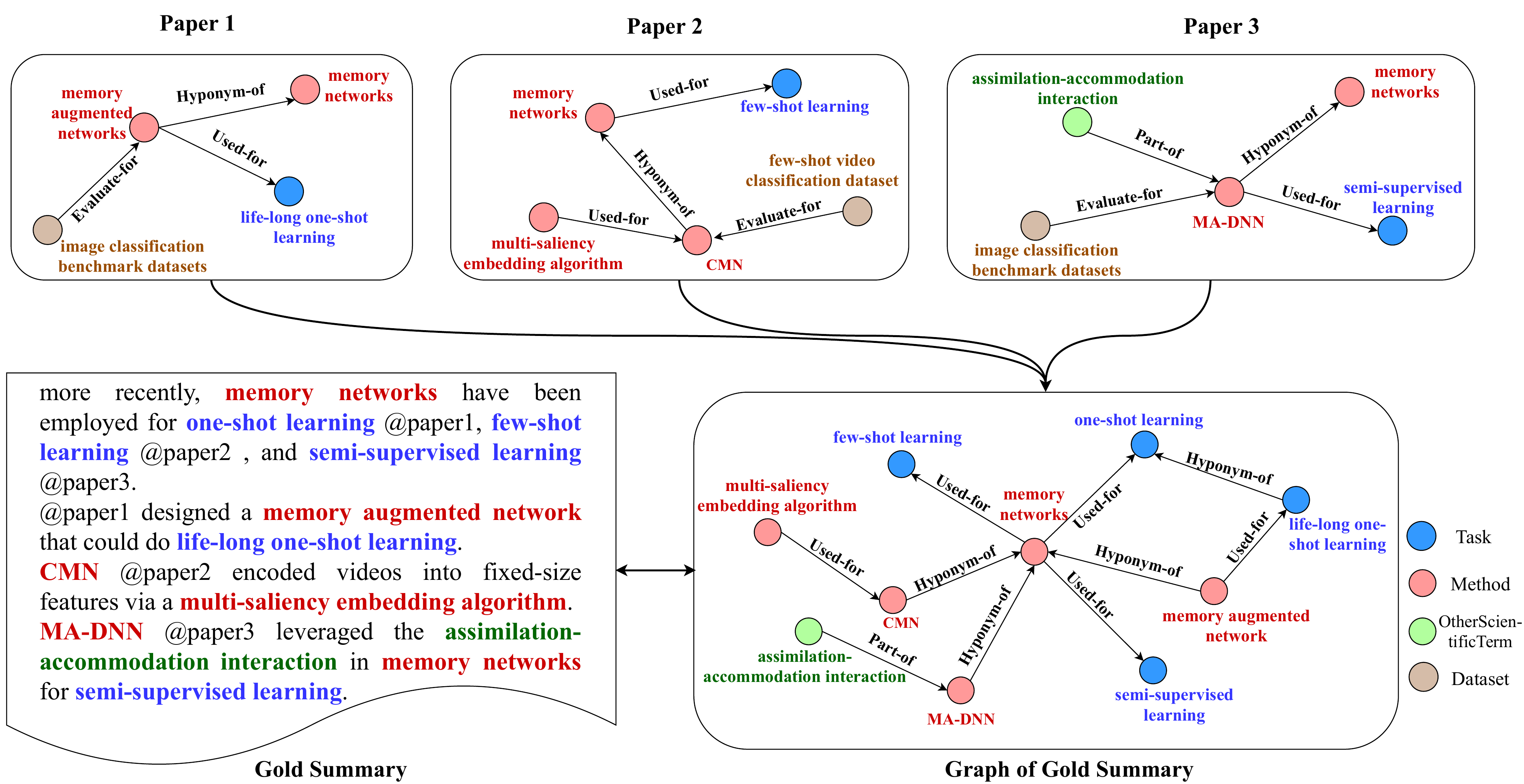}
     \caption{Knowledge graphs constructed from abstract of input scientific papers and gold summary.}
    \label{figure1}
\vspace*{-0.5\baselineskip}
\end{figure*}

To address the two aforementioned issues, we consider leveraging salient text units, namely entities, and their relations for MDSS.
Scientific papers contain multiple domain-specific entities and relations between them. These entities and relations succinctly capture important information about the main content of papers.
Knowledge graphs based on these scientific entities and relations can be inherently used for content modeling of scientific papers. Take Figure~\ref{figure1} as an example. The knowledge graph at the top left illustrates the main content of paper 1, which can be formulated as: \emph{Paper 1 uses memory augmented networks method to solve the life-long one-shot learning task, the evaluation is based on image classification benchmark datasets}. 
Furthermore, knowledge graphs can effectively capture 
cross-paper relationships through entity interactions and information aggregation. In Figure~\ref{figure1}, paper 1, 2 and 3 adopt the same method \emph{memory networks} to solve different tasks. This relationship is demonstrated in the graph of gold summary by sharing the method node \emph{memory networks}. 
\vspace*{-0.5\baselineskip}

In this paper, we develop a Transformer-based~\cite{vaswani2017attention} abstractive MDSS model, which can leverage knowledge graphs to guide paper representation and summary generation. Specifically, in the encoding part, we fuse the knowledge graphs of multiple input papers into a unified graph and design a graph updater to capture cross-paper relationships and global information. Besides, we build another graph based on the interaction between entities and sentences, and then apply an entity-sentence updater to enable information flow between nodes and update sentence representations. 

In the decoding part, knowledge graphs are utilized to guide the summary generation process via two approaches. The first is to incorporate the graph structure into the decoder by graph attention, and the second is inspired by deliberation mechanism~\citep{xia2017deliberation,li2019incremental}. Concretely, we introduce a two-stage decoder to make better use of the guidance information of knowledge graphs. The first-stage decoder concentrates on generating the knowledge graph of gold summary, while the second-stage decoder generates the summary based on the output of the first stage and the input papers. Since the knowledge graph of gold summary is in the form of graph structure, we translate the graph into equivalent descriptive sentences containing corresponding entities and relations, called \textbf{KGtext}. KGtext serves as an information-intact alternative to the knowledge graph of gold summary and is generated in the first-stage decoder, which we call the KGtext generator.

We test the effectiveness of our proposed model on Multi-XScience~\cite{lu2020multi}, a large-scale dataset for MDSS. Experimental results show that our proposed knowledge graph-centric model achieves considerable improvement compared with the baselines, indicating that knowledge graphs can exert a positive impact on paper representation and summary generation.

The main contribution is threefold: ($i$) We leverage knowledge graphs to model content of scientific papers and cross-paper relationships, and propose a novel knowledge graph-centric model for MDSS. ($ii$) We propose a two-stage decoder that introduces KGtext as intermediate output when decoding, which plays an important guiding role in the final summary generation. ($iii$) Automatic and human evaluation results on the Multi-Xscience dataset show the superiority of our model.

%% file: data/approach.tex
\section{Approach}
\label{app}

\subsection{Problem Formulation}
We first introduce the problem formulation and used notations for MDSS. Given a set of query-focused scientific papers $\mathcal{D} = \{ {d_1},{d_2},...,{d_N}\}$, where $N$ denotes the number of input papers. Each paper ${d_i}$ consists of $M_i$ sentences $\{ {s_{i,1}},{s_{i,2}},...{s_{i,{M_i}}}\}$, while each sentence $s_{i,j}$
 consists of $K_{i,j}$ words $\{ 
 {w_{i,j,1}},{w_{i,j,2}},...,{w_{i,j,{K_{i,j}}}}\}$. The gold summary $S = \{{w_1},{w_2},...{w_{{N_s}}}\}$, $N_s$ is the number of words in the gold summary. The target is to generate a summary $\hat S = \{{\hat w_1},{\hat w_2},...{\hat w_{{N_{\hat s}}}}\}$ that is close enough to the gold summary $S$.
\begin{figure*}[ht]
  \centering
   \includegraphics[width=0.95\textwidth]{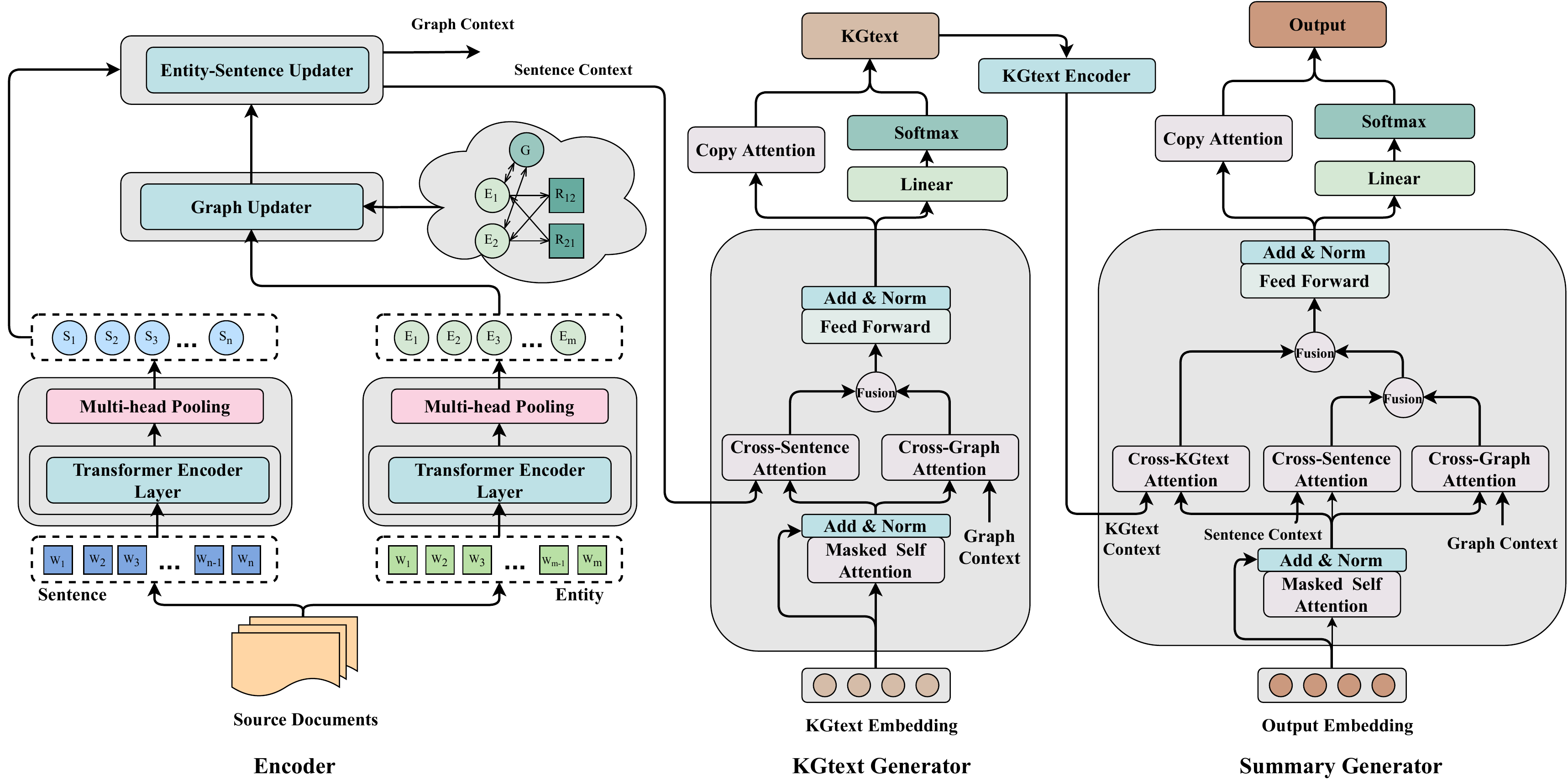}
     \caption{The overall framework of our proposed model.}
    \label{figure2}
\end{figure*}

In our two-stage decoder framework, the gold KGtext $T = \{ {w_{{t_1}}},{w_{{t_2}}},...,{w_{{t_{\hat N}}}}\}$ is also attached as input. Hence, the probability of generating the gold summary $S$ is
\begin{equation}
P(S|\mathcal{D}) = {P_{{\theta _{\mathcal{D} \to T}}}}(T|\mathcal{D})*{P_{{\theta _{(\mathcal{D},T) \to S}}}}(S|\mathcal{D},T)
\end{equation}
where ${{\theta _{\mathcal{D} \to T}}}$ and ${{\theta _{(\mathcal{D},T) \to S}}}$ are the parameters for the first-stage KGtext generator and the second-stage summary generator, respectively.

\subsection{Graph Construction}
\label{gu}
To construct the knowledge graphs for input papers, we first employ the SciIE system DYGIE++~\cite{wadden2019entity}, a well-performed science-domain information extraction system, to extract entities, relations and co-references from papers. Entities are classified into six types (\emph{Task, Method, Metric, Material, Generic}, and \emph{OtherScientificTerm}), and relations are classified into  seven types (\emph{Compare, Used-for, Feature-of, Hyponym-of, Evaluate-for, Part-of}, and \emph{Conjunction}). Besides, we collapse co-referential entity clusters into a single node based on the annotation result. 

After obtaining the knowledge graphs of multiple input papers, we fuse them into a unified graph. Then we follow the Levi transformation~\cite{levi1942} to treat each entity and relation equally. Concretely, each labeled edge is represented as two vertices: one denoting the forward relation and another denoting the reverse relation. Formally, given an entity-relation tuple $({e_1},r,{e_2})$, we create nodes ${e_1},{e_2},{r_1}$ and ${r_2}$, and add directed edges ${e_1} \to {r_1},{r_1} \to {e_2}$ and ${e_2} \to {r_2},{r_2} \to {e_1}$. In this way, the original knowledge graph is reconstructed as an unlabeled directed graph without information loss. Besides, to guarantee the connectivity of Levi graph, we add a global vertex that connects all the entity vertices. We also add entity type nodes and connect all the entities to their corresponding types.

\subsection{Model Description}
Our model follows a Transformer-based encoder-decoder architecture, shown in Figure~\ref{figure2}. The encoder includes a stack of $L_1$ token-level Transformer encoding layers to encode contextual information for tokens within each sentence and each entity. The Transformer encoding layer follows the Transformer architecture introduced in~\citet{vaswani2017attention}. The encoder also includes a \textbf{Graph Updater} to learn the graph representation of the knowledge graph and an \textbf{Entity-Sentence Updater} to update entity representation and sentence representation based on their interaction. The decoder consists of a \textbf{KGtext Generator}, which produces the descriptive sentences of the graph of gold summary, and a \textbf{Summary Generator}, which produces the final summary.

\subsection{Graph Updater}
As shown in Figure~\ref{figure2}, based on the output of token-level Transformer encoding layers, the graph updater is used to encode the knowledge graphs to obtain graph representations of input papers.

\paragraph{Node Initialization}The vertices of the constructed graph correspond to entities, relations and entity types from the SciIE annotations. Entities representations are produced using the aforementioned Transformer-based encoding method. For a given entity co-reference cluster, we first remove pronouns and stopwords and then obtain the entity representation by using the average embedding of entities in the cluster. 
For relation representation, since each relation is represented as both forward and backward vertices, we learn two embeddings per relation. We also randomly initialize the types embeddings and the global vertex embedding.

\paragraph{Contextualized Node Encoding}
We follow \citet{koncel2019text} and use a Graph Transformer to compute the hidden representations of each node in  the graph. Graph Transformer encodes each vertex ${v_i}$ using the multi-head self-attention mechanism similar to \citet{vaswani2017attention}, where each vertex representation ${{\bf{v}}_i}$ is contextualized by attending over the other vertices to which ${v_i}$ is connected in the graph. 
\begin{align}
{\hat {\bf{v}}_i} &= {{\bf{v}}_i} + \parallel _{n = 1}^N\sum\limits_{{v_j} \in {{\cal N}_i}} {\alpha _{i,j}^n{\bf{W}}_V^n} {{\bf{v}}_j}\\
\alpha _{i,j}^n &= {\rm{softmax((}}{\bf{W}}_K^n{{\bf{v}}_j}{{\rm{)}}^T}{\rm{(}}{\bf{W}}_Q^n{{\bf{v}}_i}{\rm{))}}
\end{align}
where $\parallel _{n = 1}^N$ means the concatenation of $N$ heads. ${{{\cal N}_i}}$ denotes the neighbors of $v_i$, and ${{\bf{W}}_Q^n}$, ${{\bf{W}}_K^n}$, and ${{\bf{W}}_V^n}$ are trainable parameters of query, key and value of head $n$, respectively.

\subsection{Entity-Sentence Updater} 
After getting the contextualized node embeddings for the knowledge graph, we construct an entity-sentence heterogeneous graph to update sentence representations based on the interaction between entities and sentences. The entity-sentence graph is denoted as $G = \{ V,E\}$, where $V$ stands for nodes set and $E$ stands for edges set. In the graph $G$, $V$ includes entity nodes $V_e$ and sentence nodes $V_s$, and $E$ is a real-value edge  weight matrix, where ${e_{i,j}} \ne 0$ indicates the $j$-th sentence contains the $i$-th entity.

We apply the same Graph Transformer module as the graph updater. It takes as input the entities representations from the graph updater and the sentence representations from the Transformer encoding layer, then learns the representations of nodes based on the information flow through the graph $G$. 

\subsection{KGtext Generator}
\label{bg}
\begin{figure}[t]
  \centering
   \includegraphics[width=\columnwidth]{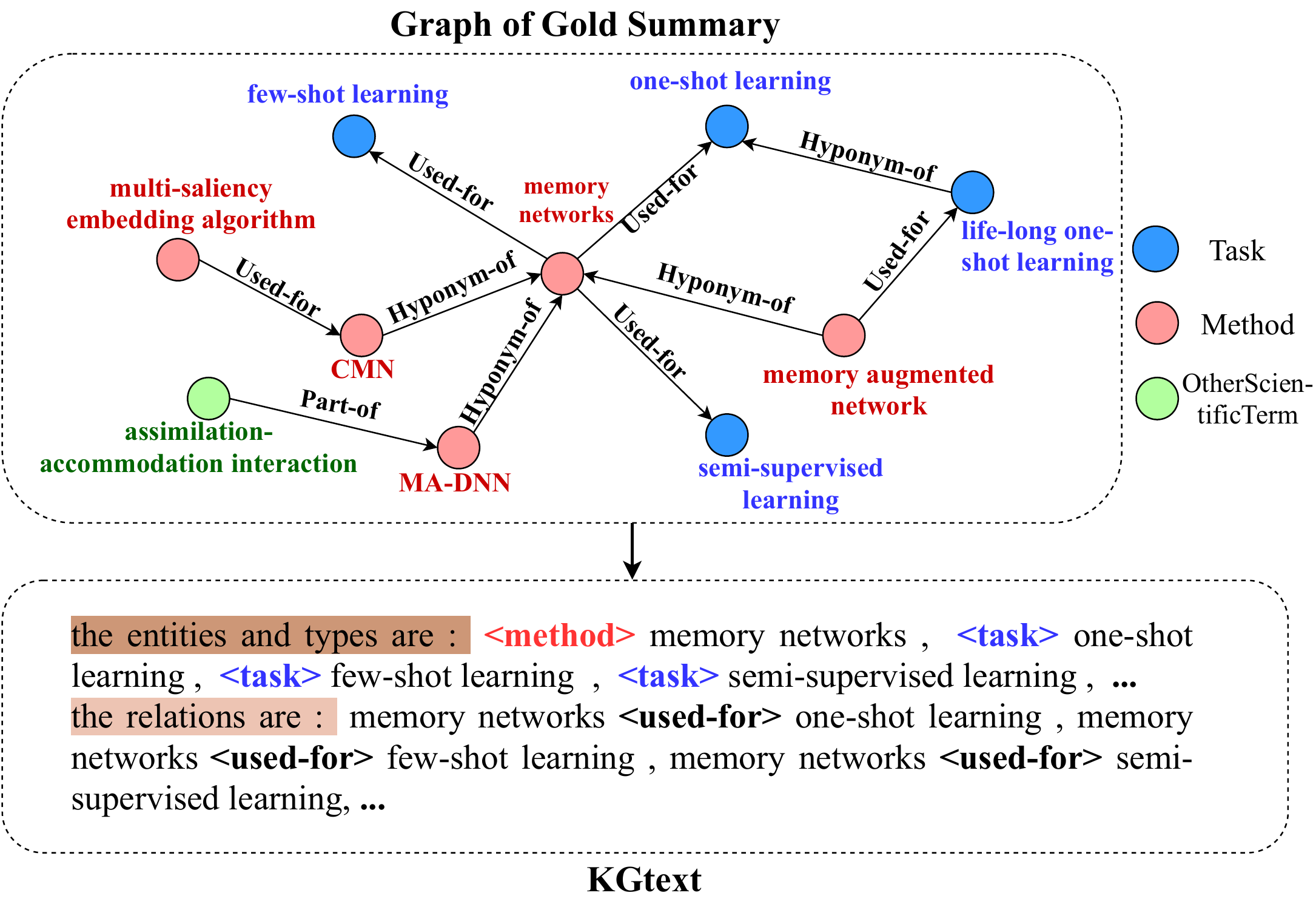}
     \caption{An example of graph of gold summary and the translated KGtext.}
    \label{figure3}
\end{figure}

In the decoding stage, we also adopt the knowledge graph-centric view and introduce the KGtext generator before the final summary generator. Here, KGtext is defined as descriptive sentences containing entities and relations translated from the knowledge graph of gold summary. An example of KGtext is shown in Figure~\ref{figure3}.

\paragraph{KGtext Construction}
To construct KGtext, we first use DYGIE++~\citep{wadden2019entity} to extract entities and relations from the human-written gold summary of the training set. Then we fill the KGtext with the prefix \emph{The entities and types are: } followed by each entity type and entity pair like \emph{<TYPE> ENT}. We also add another prompt \emph{the relations are: } to introduce the relations, in the form of \emph{ENT\_1 <REL> ENT\_2}. 

KGtext serves as an information-intact alternative to the knowledge graph of gold summary, which is generated by the KGtext generator and can provide knowledge graphs information for the final summary generation.

\paragraph{Decoding}
Since the knowledge graph of gold summary is obtained by synthesizing and simplifying the knowledge graphs of input papers via the interaction of nodes, the graph structure plays an important role in KGtext generation. Hence during decoding, we leverage source token representations as well as graph representations during KGtext decoding process. 

 We apply a stack of ${L_2}$ Transformer decoding layers as the decoder. The cross-attention sub-layer of each decoding layer computes two multi-head attention to capture both textual and graph context.
 Let $\tilde g_i^{l}$ denotes the $i$-th token output representation by the $l$-th self-attention sub-layer.
For the textual context, we use $\tilde g_i^{l}$ as query and token representations ${\bf{H_W}}$ from entity-sentence updater as keys and values.
\begin{equation}
\label{equ4}
c_{i,w}^l = {\rm{MHAtt}}(\tilde g_i^l,{{\bf{H}}_{\bf{W}}},{{\bf{H}}_{\bf{W}}})
\end{equation}
where MHAtt denotes the multi-head attention module proposed in~\citet{vaswani2017attention}. 

For the graph context, we use $\tilde g_i^{l}$ as query and entity nodes representations ${\bf{H_E}}$ from entity-sentence updater as keys and values. Considering that different entities of the input have different importance, we apply the unsupervised phrase scoring algorithm RAKE~\citep{rose2010automatic} to score the salience of entities, and incorporate entity salience into graph context computation. Given the RAKE scores $S = \{ {s_j}\}$ for entity nodes representations ${\bf{H_E}}$, we modify MHAtt module by multiplying $S$ with the attention weights.
\begin{equation}
c_{i,g}^l = {\rm{MHAtt\_Mod}}(\tilde g_i^l,{{\bf{H}}_{\bf{E}}},{{\bf{H}}_{\bf{E}}},S)
\end{equation}
where MHAtt\_Mod denotes the modified MHAtt module. And the modified attention weight $\alpha _{t}^n$ of head $n$ is calculated as 
\begin{equation}
\alpha _i^n = \frac{{{{({\rm{W}}_K^n{\bf{H_E}})}^T}({\rm{W}}_Q^n\tilde g_i^l)}}{{\sqrt {{d_{head}}} }}*S
\end{equation}
where ${{\rm{W}}_K^n}$ and ${{\rm{W}}_Q^n}$ are parameter weights, $d_{head}$ denotes the dimension of each attention head.

We then add a fusion gate to merge both the textual context and the graph context.
\begin{align}
{c_i^l} &= z*c_{i,w}^l + (1 - z)*c_{i,g}^l\\
z &= {\rm{sigmoid}}([c_{i,w}^l;c_{i,g}^l]{{\rm{W}}_f} + {b_f})
\end{align}
where ${\rm{W}}_f$ and $b_f$ are the linear transformation parameter. The feed-forward network is used to further transform the output.
\begin{equation}
g_i^l = {\rm{LayerNorm}}(c_i^l + {\rm{FFN}}(c_i^l))
\end{equation}

The generation distribution $p_t^g$ over the target vocabulary is calculated by feeding the output $g_t^{{L_2}}$ to a softmax layer.
\begin{equation}
p_i^g = {\rm{softmax(}}g_i^{{L_2}}{{\rm{W}}_g}{\rm{ + }}{b_g}{\rm{)}}
\end{equation}
where ${{\rm{W}}_g}$ and ${b_g}$ are learnable linear transformation parameter.

Furthermore, we also employ the copy mechanism~\citep{see2017get} to alleviate the out-of-vocabulary (OOV) problem. The final generation distribution $p_i^t$ is the "mixture" of both $p_i^g$ and the copy probability over source words $p_i^c$.

The loss is the negative log likelihood of the gold KGtext $w_{{t_i}}$:
\begin{equation}
{L_T} =  - \frac{1}{{\hat N}}\sum\nolimits_{i = 1}^{\hat N} {\log p_i^t({w_{{t_i}}})} 
\end{equation}
\subsection{Summary Generator}
The final summary generator has a similar decoding architecture to the KGtext generator, but differs in that the summary generator utilizes the generated KGtext to guide summary  generation. 

Given the KGtext generative distribution $\{p_i^t\}$, we obtain the decoding sequence of KGtext $\hat T$ by greedy search during training. Then we add an encoder similar to the aforementioned sentence encoder to get the KGtext representations ${{\bf{H}}_{\bf{T}}}$. Besides attending to textual and graph context,  we use the same multi-head attention as equation~(\ref{equ4}) to compute KGtext context $\hat c_{i,t}^l$ to capture KGtext influence. 

Together with the textual context $\hat c_{i,w}^l$ and the graph context $\hat c_{i,g}^l$, we apply a hierarchical fusion mechanism to combine the three contexts, by first merging the textual context and the graph context, and then the KGtext context.
\begin{align}
\hat c_i^l &= {z_1}*\hat c_{i,w}^l + (1 - {z_1})*\hat c_{i,g}^l\\
{z_1} &= {\rm{sigmoid}}([\hat c_{i,w}^l;\hat c_{i,g}^l]{{\rm{W}}_{1,f}} + {b_{1,f}})\\
c_i^l &= {z_2}*\hat c_i^l + (1 - {z_2})*\hat c_{i,t}^l\\
{z_2} &= {\rm{sigmoid}}([\hat c_i^l;\hat c_{i,t}^l]{{\rm{W}}_{2,f}} + {b_{2,f}})
\end{align}
where ${\rm{W}}_{1,f}$, $b_{1,f}$, ${\rm{W}}_{2,f}$ and $b_{2,f}$ are the linear transformation parameter.

Given the final summary generation distribution $p_i^s$, the loss is the negative log likelihood of the gold summary $w_i$:
\begin{equation}
{L_S} =  - \frac{1}{{{N_s}}}\sum\nolimits_{i = 1}^{{N_s}} {\log p_i^s({w_i})} 
\end{equation}

\subsection{Training Strategy}
We train the KGtext generator and the summary generator in a unified architecture in an end-to-end manner. Furthermore, in practice, we find the KGtext generated from greedy search has a strong influence on the summary generation. The low-quality KGtext greatly impairs the performance of the model. Hence, we train another auxiliary decoder on top of ${P_{{\theta _{\mathcal{D} \to {\rm{S}}}}}}(S|\mathcal{D})$, which uses a one-stage decoder without generating KGtext. It has the same architecture as the summary generator except for the cross-attention on KGtext.

Given the final summary generation distribution of the auxiliary decoder $\tilde p_i^s$, the loss is the negative log likelihood of the gold summary $w_i$:
\begin{equation}
{L_A} =  - \frac{1}{{{N_s}}}\sum\nolimits_{i = 1}^{{N_s}} {\log \tilde p_i^s({w_i})} 
\end{equation}

The final loss function is:
\begin{equation}
{\cal L} = {{\cal L}_S} + \lambda {{\cal L}_T} + \eta {{\cal L}_A}
\end{equation}
where $\lambda$ and $\eta$ are both hyper parameters. In this way, we can reduce the effect of some low-quality generated KGtext and improve the stability of our model.

%% file: data/experiments.tex
\section{Experiments}
\subsection{Dataset}
We conduct experiments on the recently released Multi-Xscience dataset~\citep{lu2020multi}, which is the first large-scale and open MDSS dataset. It contains 30,369 instances for training, 5,066 for validation and 5,093 for test. On average, each source paper cluster contains 4.42 papers and 778.08 words, and each gold summary contains 116.44 words.

\subsection{Implementation Details}
We set our model parameters based on preliminary experiments on the validation set. We prune the vocabulary to 50K. The number of encoding layer $L_1$ and the number of decoding layer $L_2$ are both 6. We set the dimension of word embeddings and hidden size to 256, feed-forward size to 1,024. We set 8 heads for multi-head attention. For the Graph Transformer of the graph updater and the entity-sentence updater, we set the number of iterations to 3. We set dropout rate to 0.1 and label smoothing~\citep{szegedy2016rethinking} factor to 0.1. We use Adam optimizer with learning rate $\alpha=0.02$, $\beta_1=0.9$, $\beta_2=0.998$; we also apply learning rate warmup over the first 8000 steps,  and decay as in~\citet{vaswani2017attention}. The batch size is set to 8. $\lambda$ and $\eta$ are both set to 1.0. The model is trained on 1 GPU (NVIDIA Tesla V100, 32G) for 100,000 steps. We select the top-3 best checkpoints based on performance on the validation set and report averaged results on the test set.

For KGtext decoding, we use greedy search with the minimal generation length 100, while for summary decoding, we use beam search with beam  size 5 and the minimal generation length is 110, consistent with~\citet{lu2020multi}. The length penalty factor is set to 0.4.

\subsection{Metrics and Baselines}
We use ROUGE $F_1$~\citep{lin2004rouge} to evaluate the summarization quality. Following previous work, we report unigram and bigram overlap (ROUGE-1 and ROUGE-2) as a means of assessing informativeness and the longest common subsequence (ROUGE-L) as a means of assessing fluency. 

We compare our model with several typical extractive and abstractive summarization models. Due to space limitations, we put the introduction of these models in appendix~\ref{appen1}.

\subsection{Automatic Evaluation}
\begin{table}[tb]
    \centering
     \resizebox{\columnwidth}{!}{
    \hfill
    \renewcommand\arraystretch{1.2}
    \begin{tabular}{l|ccc}
    \hline
    \textbf{Model} & \textbf{R-1} & \textbf{R-2} & \textbf{R-L} \\ \hline
    \multicolumn{4}{l}{\emph{Extractive}}\\ \cline{1-4}
    LexRank~\citep{erkan2004lexrank} & 30.19& 5.53& 26.19\\
    TextRank~\citep{mihalcea2004textrank} & 31.51& 5.83& 26.58\\
    HeterSumGraph$^*$~\cite{wang2020heterogeneous} & 31.36& 5.82& 27.41\\
    Ext-Oracle & 38.45 & 9.93 & 27.11 \\\hline
    \multicolumn{4}{l}{\emph{Abstractive}}\\\hline
    GraphSum$^*$~\citep{li2020leveraging} & 29.58 & 5.54 & 26.52 \\
    Hiersumm~\citep{liu2019hierarchical} & 30.02& 5.04& 27.6\\
    HiMAP~\citep{fabbri2019multi} & 31.66& 5.91& 28.43\\
    BertABS~\citep{liu2019text} & 31.56& 5.02& 28.05\\
    BART~\citep{lewis2020bart} & 32.83& 6.36& 26.61\\
    SciBertABS~\citep{lu2020multi} & 32.12& 5.59& 29.01\\
    MGSum$^*$~\citep{jin2020multi} & 33.11& 6.75& 29.43\\
    Pointer-Generator~\citep{see2017get} & 34.11& 6.76& 30.63\\
    \hline
    KGSum & \textbf{35.77}& \textbf{7.51}&  \textbf{31.43} \\
    
    \hline
    \end{tabular}
    }
    \caption{ROUGE F1 evaluation results on the test set of Multi-Xscience. The results marked with * are obtained by running the released code using the same beam size and decoding length. Other results without * are from ~\citet{lu2020multi}.}
    \label{table1}
\end{table}

Table~\ref{table1} summarizes the evaluation results on the Multi-Xscience dataset. 

As can be seen, abstractive models generally outperform extractive models, especially on ROUGE-L, showing that abstractive models can generate more fluent summaries. 
Among the abstractive baselines, Pointer-Generator~\citep{see2017get} surprisingly outperforms other newer models. We partially attribute this result to Pointer-Generator designing an additional coverage mechanism~\citep{tu2016modeling} to effectively reduce redundancy. This result also illustrates that MDSS is challenging and requires domain-specific solutions for paper content representation and cross-paper relationship modeling.

The last block reports the result of our model KGSum. KGSum outperforms any other models, achieving scores of 35.77, 7.51, and 31.43 on the three ROUGE metrics. Our model surpasses other models by a remarkable large margin (at least improving 1.66\%, 0.75\%, and 0.80\%). The result demonstrates that our model can generate more informative and more coherent summaries, indicating
the effectiveness of our proposed knowledge graph-centric encoder and decoder framework.

\begin{table}[tb]
    \centering
    \resizebox{\columnwidth}{!}{
    \hfill
     \renewcommand\arraystretch{1.2}
    \begin{tabular}{l|llll}
    \hline
    \textbf{Model} & \textbf{Overall} & \textbf{Inf} & \textbf{Fluency} & \textbf{Succ} \\ \hline
    GraphSum & -1.42$^*$& -1.47$^*$&  -1.08$^*$ & -1.23$^*$\\
    MGSum & -0.38$^*$& 0.60& -0.20$^*$ & -0.55$^*$\\
    Pointer-Generator & 0.62$^*$& 0.31$^*$& 0.17$^*$ & 0.60$^*$\\
    KGSum & \textbf{1.30} & \textbf{0.68} & \textbf{1.17} & \textbf{1.22} \\
    \hline
    \end{tabular}
    }
    \caption{Human evaluation of system summaries on Multi-Xscience test set.  Inf stands for \emph{informativeness} and Succ stands for \emph{succinctness}. The larger rating denotes better summary quality. * indicates the ratings of the corresponding model are significantly (by Welch's t-test with $p < 0.05$) outperformed by our model. The inter-annotator agreement score (Fleiss Kappa) is 0.63, which indicates substantial agreement between annotators.}
    \label{table5}
\end{table}

\begin{table}[tb]
    \centering
    \resizebox{\columnwidth}{!}{
    \hfill
    \small
     \renewcommand\arraystretch{1.2}
    \begin{tabular}{l|lll}
    \hline
    \textbf{Model} & \textbf{R-1} & \textbf{R-2} & \textbf{R-L} \\ \hline
    KGSum & \textbf{35.77}& \textbf{7.51}&  \textbf{31.43}\\
    - KGG & 35.34& 7.28& 30.91\\
    - KGG - RAKE & 35.17& 7.18& 30.75\\
    - KGG - RAKE - GU & 34.97 & 7.08 & 30.63 \\
    - KGG - RAKE - GU - ESU & 34.79 & 6.90 & 30.36 \\
    \hline
    \end{tabular}
    }
    \caption{Ablation studies on Multi-Xscience test set. We remove various modules and explore their influence on our model. '-' means the removal operation from KGSum. The last row (-KGG-RAKE-GU-ESU) is the clean baseline without any module we propose. }
    \label{table2}
\end{table}

\subsection{Human Evaluation}
We further access the linguistic quality of generated summaries by human evaluation. We randomly select 30 test instances from the Multi-Xscience test set, and invite three graduate students as annotators to evaluate the outputs of different models independently. Annotators assess the overall quality of summaries by ranking them considering the following criteria:: (1) \emph{Informativeness}: does the summary convey important facts of the input papers? (2) \emph{Fluency}: is the summary fluent and grammatical? (3) \emph{Succinctness}: whether the summary contains repeated information?
Annotators are asked to rank all systems from 1 (best) to 4 (worst). All systems get score 2, 1, -1, -2 for ranking 1, 2, 3, 4 respectively. The rating of each system is computed by averaging the scores on all test instances. 

The result is shown in Table~\ref{table5}. The overall rating and the ratings for the above three aspects are reported. We can see that KGSum performs much better than other models. The overall rating of KGSum achieves 1.2, which is much higher than 0.62, -0.38, and -1.42 of Pointer-Generator, MGSum, and GraphSum. The results on informativeness indicate our model can effectively capture the salient information of papers and generate more informative summaries.
The results on fluency and succinctness indicate that KGSum is able to generate more fluent and concise summaries. 
Furthermore, Pointer-Generator achieves a much higher score on succinctness than MGSum, which further proves that Pointer-Generator generates less redundant summaries and thus has better performance.

\begin{figure}[t]  
    \centering   
    
    \subfloat[Recall of EW] 
    {
        \begin{minipage}[t]{0.5\columnwidth}
            \centering        
         \includegraphics[width=\columnwidth]{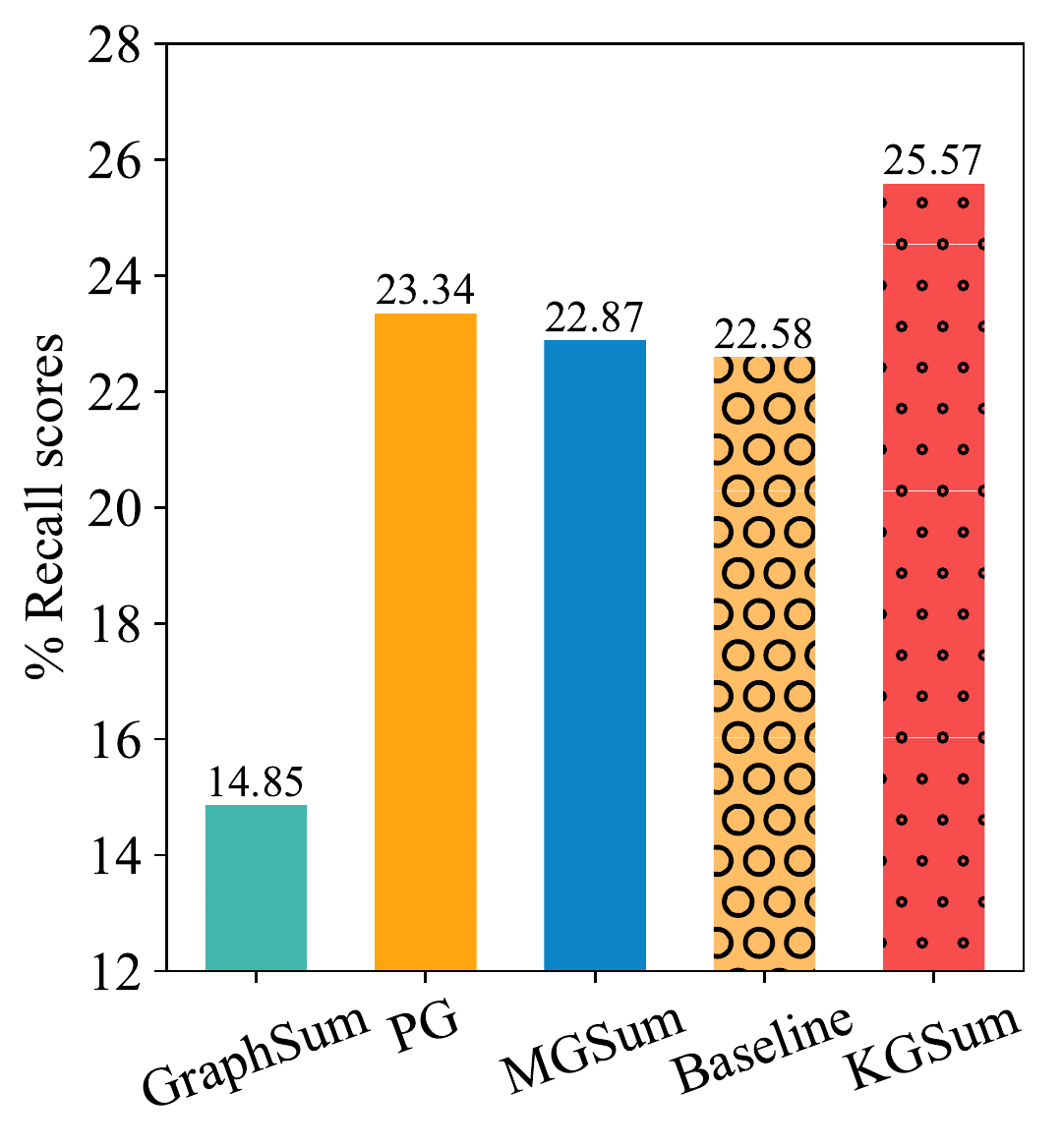} 
        \end{minipage}%
    }
    \subfloat[Recall of CW] 
    {
        \begin{minipage}[t]{0.5\columnwidth}
            \centering     
            \includegraphics[width=\columnwidth]{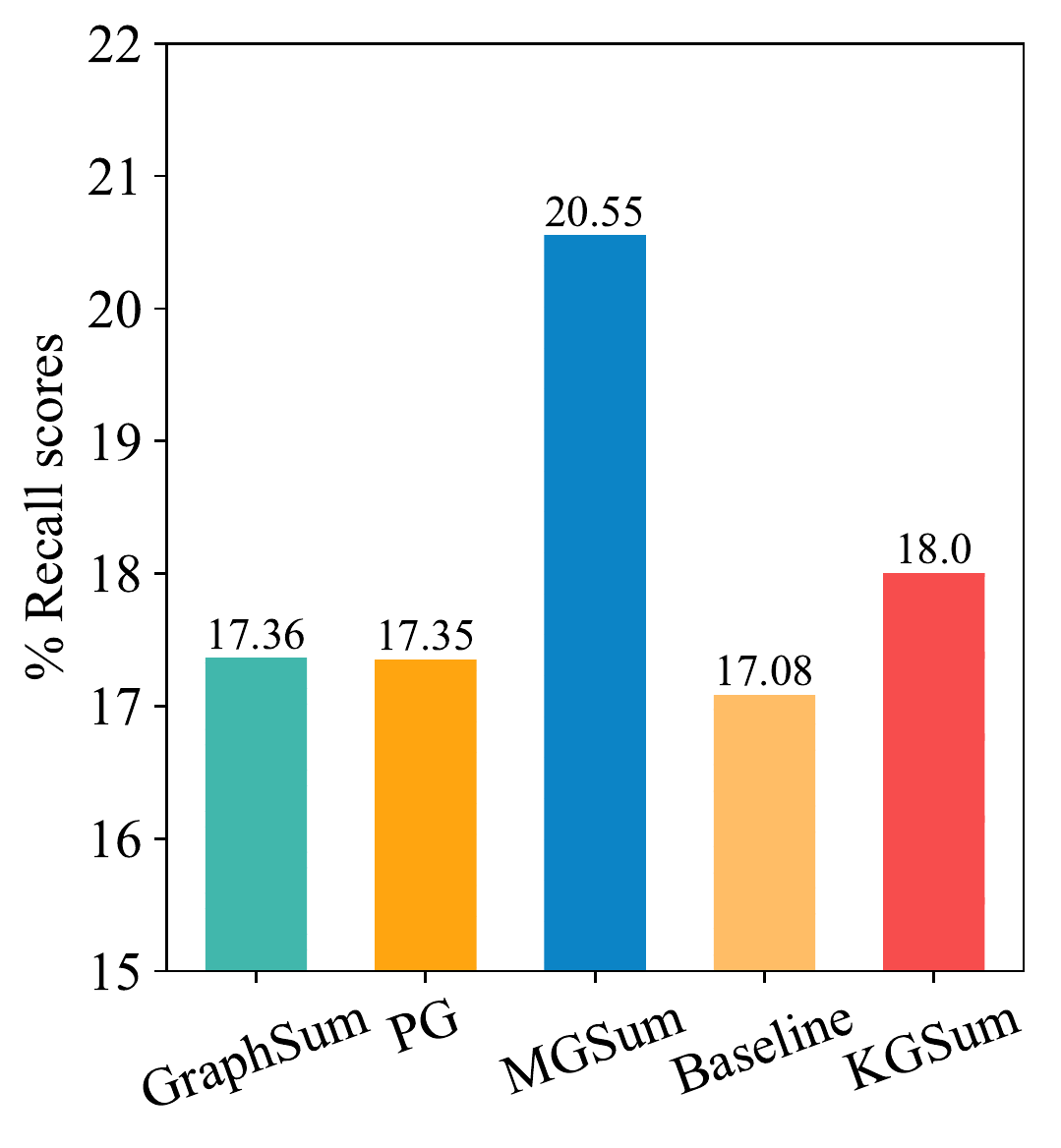} 
        \end{minipage}
        \label{fig:b} 
    }%
    
    \caption{Recall of EW and CW for different models on Multi-Xscience test set. PG stands for Pointer-Generator, Baseline is our Transformer-based model without any module we propose.} 
    \label{figure7} 
\end{figure}

\subsection{Model Analysis}
For a thorough understanding of KGSum, we conduct several experiments on Multi-Xscience test set.

\paragraph{Ablation Study} Firstly, we perform an ablation study to validate the effectiveness of individual components. Here, \textbf{KGG} stands for KGtext generator, \textbf{RAKE} refers to the RAKE algorithm that measures entity salience, \textbf{GU} stands for graph updater, \textbf{ESU} stands for entity-sentence updater. We remove KGG, RAKE, GU, ESU one by one in order from decoder to encoder. The result is illustrated in Table~\ref{table2}. We find that the GU and ESU module in the encoder can effectively encode knowledge graph information and utilize knowledge graphs to enable better information flowing between text nodes, which is conducive to content modeling and relationship modeling. Using RAKE to measure entity salience also benefits a lot for graph context computation when decoding. Further, the KGG module also brings significant improvement, indicating our proposed two-stage decoder with KGtext generator is effective in generating summary under the guidance of knowledge graphs.

\paragraph{Recall of Entity Words} In order to intuitively demonstrate the impact of knowledge graphs, we investigate the recall of gold summary entities in the generated summary.
The exact match of entities is difficult because entities have different mentions. Therefore, we count recall of entity words instead.
We classify the words in papers into two categories: \textbf{Entity Words} (EW) and \textbf{Context Words} (CW). EW are defined as words in papers that are recognized as entities by SciIE tools, while CW are words other than EW. We exclude stopwords when calculating EW and CW, because stopwords have no practical meaning. Then, we define \emph{Recall of EW} as:
\begin{equation}
\small
Recal{l_{EW}} = \frac{{\sum\limits_{S \in \{ Ref\} } {\sum\limits_{{N_{EW}} \in S} {Coun{t_{match}}({N_{EW}})} } }}{{\sum\limits_{S \in \{ Ref\} } {\sum\limits_{{N_{EW}} \in S} {Count({N_{EW}})} } }}
\end{equation}
where ${\{ Ref\} }$ denotes the gold summaries, ${Coun{t_{match}}({N_{EW}})}$ denotes the number of overlapped EW in the gold summaries and  the generated summaries. ${Count({N_{EW}})}$ denotes the number of EW in the gold summaries. \emph{Recall of CW} is defined in a similar manner. 

The results are shown in Figure~\ref{figure7}. We find that KGSum achieves the highest recall of EW, compared with the baseline model and other models. The result proves that our model focuses on more entity information under the guidance of knowledge graphs. Conversely, in Figure~\ref{fig:b}, MGSum achieves the highest recall of CW, but ROUGE-1/2/L scores of MGSum are only 33.11/6.75/29.43, falling behind KGSum. The result indicates that recall of CW has limited effect on model performance, which is in line with our intuition since CW do not contain important semantic information. 
\begin{table}[tb]
    \centering
    \resizebox{0.95\columnwidth}{!}{
    \hfill
    \small
     \renewcommand\arraystretch{1.2}
    \begin{tabular}{l|lll}
    \hline
    \textbf{KGtext Variants} & \textbf{R-1} & \textbf{R-2} & \textbf{R-L} \\ \hline
    Ent & 35.61 & 7.43 & 31.24 \\
    Ent+Type & 35.67 & 7.42 & 31.29 \\
    Ent+Type+Rel & \textbf{35.77}& \textbf{7.51}&  \textbf{31.43}\\
    \hline
    \end{tabular}
    }
    \caption{Analysis of the impact of different KGtext contents on summarization.}
    \label{table4}
\end{table}

\paragraph{Influence of KGtext Contents} We also conduct experiments to analyze the impact of different KGtext contents on MDSS. We consider the following three variants: (1) only entities (Ent), (2) entities + types (Ent+Type), (3) entities + types + relations (Ent+Type+Rel), to construct the KGtext using the same strategy in section~\ref{bg}. Result in Table~\ref{table4} demonstrates MDSS can benefit from 
different components of knowledge graph, including entities, types and relations.

%% file: data/relatedwork.tex
\section{Related Work}
Early MDSS works are mainly based on artificially constructed small-scale datasets, using unsupervised extractive methods to extract sentences from multiple papers. \citet{mohammad2009using} take citation texts, paper abstracts and full paper texts as input for survey generation. They conduct the experiment with just two instances. \citet{jha2015surveyor} construct 15 instances and combine a content model with a discourse model to generate coherent scientific summarizations. \citet{hoang2010towards} construct 20 instances, each with an annotated topic hierarchy tree, to generate summarization for multiple scientific papers. Similar works also exist in~\citep{jha2015content,hu2014automatic,yang2017keyphraseds}. These unsupervised works are crude in both content modeling and relationship modeling and fail to generate high-quality summaries.

Some subsequent efforts apply deep learning-based methods to MDSS using large-scale datasets ~\citep{wang2018neural,jiang2019hsds,chen2021capturing}.~\citet{wang2018neural} build a dataset with 8080 instances and construct a heterogeneous bibliography graph, and then utilize a CNN and RNN-based model for extractive summarization.~\citet{jiang2019hsds} collect 390,000 instances, and 
use a hierarchical encoder and a two-step decoder to generate summary in an abstractive manner for the first time.~\citet{chen2021capturing} collect two large-scale datasets with 136,655 and 80,927 instances, respectively. They apply a Transformer-based model to capture the relevance between papers for abstractive summarization. However, all the above works neglect salient semantic units to  capture semantic information and relationships between papers. In this paper, based on Mutli-Xscience~\citep{lu2020multi}, we use knowledge graph information to model content and relationships between papers.

%% file: data/conclusion.tex
\section{Conclusion}
In this work, we propose a knowledge graph-centric Transformer-based model for MDSS. Our model is able to incorporate knowledge graph information into the paper encoding process with a graph updater and an entity-sentence updater, and introduce a two-stage decoder including a KGtext generator and a summary generator to guide the summary decoding with knowledge graph information. Experiments show that the proposed model significantly outperforms all strong baselines and achieves the best result on the Multi-Xscience dataset.

In the future, we will explore other more intuitive and effective methods to incorporate graph information in both the encoding and decoding phase of summary generation.